%% file: main.tex
\documentclass[10pt,twocolumn,letterpaper]{article}

\usepackage[pagenumbers]{cvpr}
\usepackage{graphicx}
\usepackage{amsmath}
\usepackage{amssymb}
\usepackage{booktabs}
\usepackage{multirow}
\usepackage{soul}
\usepackage{marvosym}
\usepackage{inconsolata}

\usepackage[toc,page]{appendix}
\usepackage[pagebackref,breaklinks,colorlinks]{hyperref}
\usepackage[capitalize]{cleveref}
\crefname{section}{Sec.}{Secs.}
\Crefname{section}{Section}{Sections}
\Crefname{table}{Table}{Tables}
\crefname{table}{Tab.}{Tabs.}

\begin{document}

\title{Constraint and Union for Partially-Supervised Temporal Sentence Grounding}

\author{Chen Ju\textsuperscript{1$\ast$}, \ \ Haicheng Wang\textsuperscript{1$\ast$}, \ \ Jinxiang Liu\textsuperscript{1}, \ \ Chaofan Ma\textsuperscript{1}, \ \  Ya Zhang\textsuperscript{1\Letter}, \\
Peisen Zhao\textsuperscript{2}, \ \ Jianlong Chang\textsuperscript{2}, \ \ Qi Tian\textsuperscript{2} 
\and
\textsuperscript{1}{CMIC, Shanghai Jiao Tong University} \ \  \textsuperscript{2}{Huawei Cloud \& AI} \\
{\tt\small \{ju\_chen,\;anakin\_skywalker,\;jinxliu,\;chaofanma,\;ya\_zhang\}@sjtu.edu.cn} 
\\
{\tt\small pszhao93@gmail.com} \ \
{\tt\small \{jianlong.chang,\;tian.qi1\}@huawei.com}
}

\maketitle

\begin{abstract}
Temporal sentence grounding aims to detect the event timestamps described by the natural language query from given untrimmed videos. The existing fully-supervised setting achieves great performance but requires expensive annotation costs; while the weakly-supervised setting adopts cheap labels but performs poorly. To pursue high performance with less annotation cost, this paper introduces an intermediate partially-supervised setting, i.e., only short-clip or even single-frame labels are available during training. To take full advantage of partial labels, we propose a novel quadruple constraint pipeline to comprehensively shape event-query aligned representations, covering intra- and inter-samples, uni- and multi-modalities. The former raises intra-cluster compactness and inter-cluster separability; while the latter enables event-background separation and event-query gather. To achieve more powerful performance with explicit grounding optimization, we further introduce a partial-full union framework, i.e., bridging with an additional fully-supervised branch, to enjoy its impressive grounding bonus, and be robust to partial annotations. Extensive experiments and ablations on Charades-STA and ActivityNet Captions demonstrate the significance of partial supervision and our superior performance.
\end{abstract}

\section{Introduction}
Temporal sentence grounding (TSG) plays an important role for video-language understanding. The goal is to detect the start and end timestamps of the event, described by a given natural language query from untrimmed videos. TSG exists extensive application scenarios~\cite{yao2016highlight,shu2015joint}, and serves as the key for multi-modal representation learning.

TSG has recently developed two widely-used settings. One is the fully-supervised setting (FTSG)~\cite{gao2017tall,anne2017localizing,liu2018cross}, {\em i.e.}, each event is annotated with precise temporal boundaries. Although the great progress has been made in this research line, labeling precise event boundaries is time-consuming and quite subjective. Because for events with high-level semantic concepts, it is difficult to strictly identify the identical event boundaries for different annotators. To significantly alleviate annotation burdens, the weakly-supervised setting (WTSG) has also been extensively studied~\cite{mithun2019weakly,lin2020weakly}, {\em i.e.}, during training, only text queries are provided for each untrimmed video without any temporal annotations. However, solely digging from such a barren supervision fundamentally limits the empirical performance, thus leaving a large margin from applications in reality.

\begin{figure}[t]
\begin{center}
\includegraphics[width=0.47\textwidth] {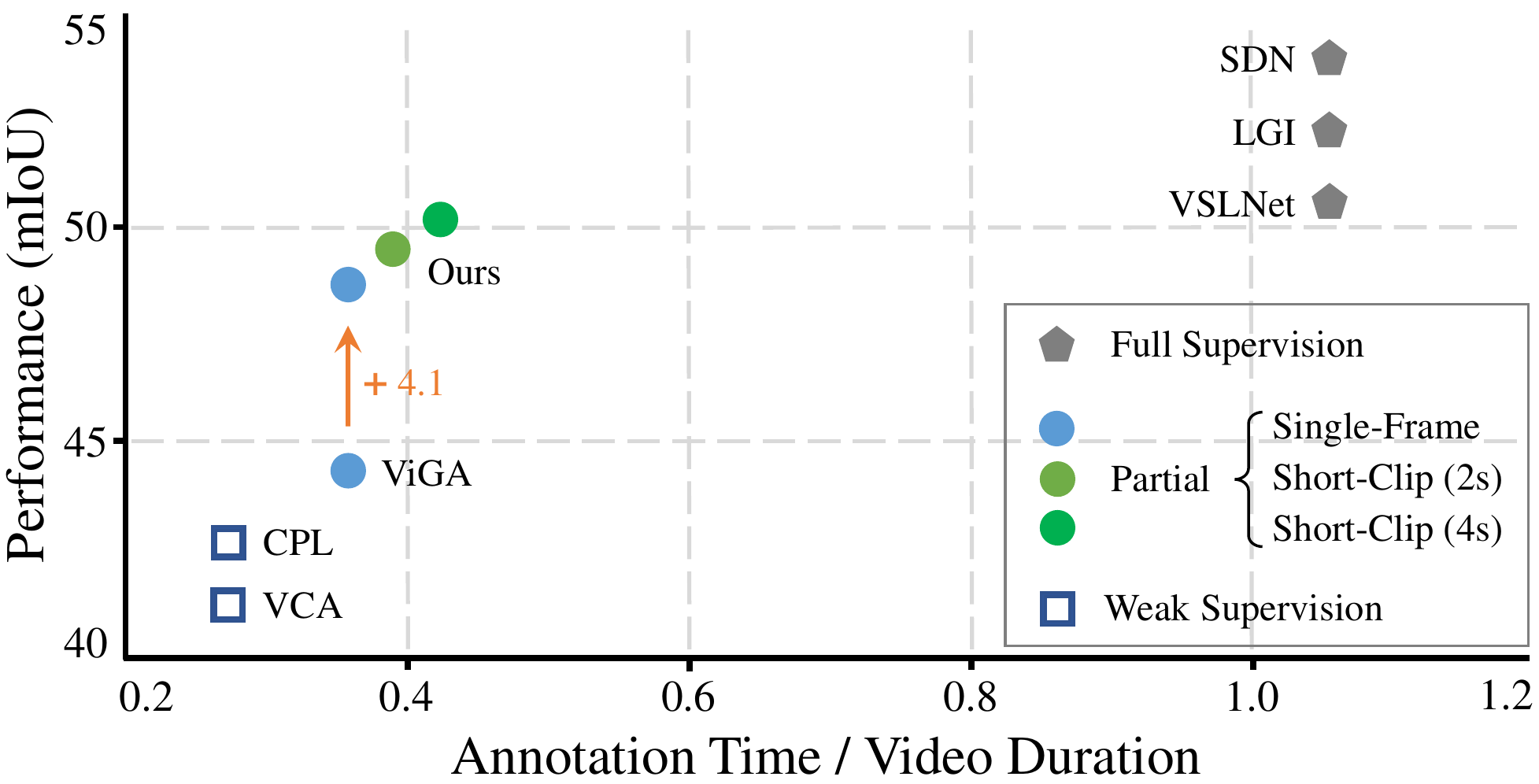}
\end{center}
\vspace{-0.5cm}
\caption{\textbf{Performance {\em vs.} Annotation Costs.} For temporal sentence grounding, full supervision achieves good performance, but requires expensive annotation; weak supervision significantly alleviates annotation costs, but shows poor performance. To achieve great empirical performance and maintain cheap annotation overhead at the same time, we introduce the intermediate setting: partial supervision, including short-clip or even single-frame labels.}
\label{fig:intro}
\end{figure}

In this situation, one question naturally arises: {\em is there an intermediate setting between full and weak supervisions, which obtains relatively high performance but requires less annotation cost?} This paper answers the question by introducing a {\em partially-supervised} setting (PTSG), inspired by scribble supervision~\cite{lin2016scribblesup,bearman2016s} of the image domain. Specifically, for each text query, a partial temporal region, corresponding to a short video-clip, is randomly annotated within the whole event interval. Note that, in the strictest case, partial labels will degenerate to single-frame labels~\cite{cui2022video}, {\em i.e.}, labeling one timestamp for each event. Figure~\ref{fig:intro} compares the performance in terms of annotation costs between various supervisions. Such partial supervision greatly improves the grounding performance, at the cost of a slight increase in annotation time, which is very effective.

Hereafter, our goal is to detect complete event intervals through limited yet precise partial labels. We design from: multi-modal representations and grounding framework.

To achieve high-quality representations, one trivial solution~\cite{cui2022video} is to construct contrastive learning by treating labeled frames and paired queries as positive, while non-pairs as negative. However, it suffers from two evident issues. (1) Partially annotated frames are semantically incomplete comparing to the whole events; (2) Relationships between events (or samples) are not fully exploited.

To solve these issues, we propose one novel quadruple constraint pipeline for PTSG. The first two constraints are built on intra-samples. They promote event-query gather for multi-modal correspondence and raise event-background separation for the visual uni-modality. To obtain complete event intervals, we introduce an event detector which takes the partial labels as seed anchors and extends them for an event mask. Then, features for event and background can be calculated via the event mask. To build more semantic constraints from the whole dataset, another two constraints are proposed for inter-samples. The events with similar text queries are clustered, given that TSG has no intuitive category. Then, constraint losses are applied to further enable intra-cluster compactness and inter-cluster separability.

For grounding results, one vanilla solution is to use the output from the event detector. But partial labels (grounding anchors) are unavailable during inference, and lacking explicit grounding objectives causes excessively noisy results. We here solve the issues via a simple yet effective framework: partial-full union (PFU). Specifically, we bridge the partially-supervised branch with a fully-supervised branch by grounding pseudo-labels, turning PTSG into FTSG.

This PFU framework can do more at one stroke. Structurally, it bridges the gap between full and partial supervisions, enabling PTSG to enjoy strong grounding bonuses from fully-supervised methods, {\em e.g.}, explicit grounding optimization, superior network architecture~\cite{li2021proposal} and detection paradigm~\cite{li2022compositional}. Functionally, it plays a similar role as self-training. The fully-supervised branch could refine and denoise pseudo-labels of the partially-supervised branch for better grounding. From the perspective of annotations, partial labels have a high degree of freedom in event intervals, posing a great challenge to framework robustness. Utilizing the full-supervised branch, our framework obtains consistent effectiveness to various label distributions, also avoids the labeling gap between training and testing in PTSG.

On two public datasets: Charades-STA and ActivityNet Captions, we annotate partial labels and show their significance. Our method shows superior performance over competitors. We further perform extensive ablations to dissect each component, both quantitatively and qualitatively.

To sum up, our contributions lie in three folds:

$\bullet$ We propose the partial supervision for TSG, and introduce a partial-full union framework to bridge FTSG and PTSG, thus enjoying strong grounding bonuses from fully-supervised methods, and being robust to partial labels.

$\bullet$ We propose one novel and comprehensive quadruple constraint pipeline, including uni- and multi-modality constraints, intra- and inter-sample constraints, to align event-query representations with high-quality.

$\bullet$ We conduct extensive experiments and ablation studies to show the significance of partial supervision, and our superior performance on public benchmarks.

\section{Related Work}
\noindent {\bf Fully-supervised Temporal Sentence Grounding} gives start and end timestamps for each query during training~\cite{gao2017tall,anne2017localizing}. Based on boundary labels, many special designs are made for advanced grounding frameworks: powerful architectures and explicit grounding optimization. The proposal-based framework first obtains event candidates by sliding windows~\cite{liu2018attentive,liu2018cross}, pre-defined anchors~\cite{chen2018temporally,cao2021pursuit,lin2020moment,liu2022memory,liu2021context,liu2020jointly,liu2021progressively,liu2022exploring,zhang2019man,zhang2019cross}, or learnable networks~\cite{chen2019semantic,liu2021adaptive,xiao2021boundary,xu2019multilevel}, then ranks proposals to match queries. The proposal-free framework interacts each frame with queries, then directly predicts results. Specifically, regression-based methods~\cite{chen2020learning,chen2020hierarchical,chen2021end,lu2019debug,yuan2019find} regress boundaries for refinement, while span-based methods~\cite{chen2020rethinking,hao2022query,hou2021conquer,nan2021interventional,rodriguez2020proposal,zhang2021natural,zhao2020bottom,ju2022prompting} identify frame-wise boundary probabilities. Nevertheless, all above methods demand expensive boundary labels, widening model differences between full and weak supervisions. This paper uses the PFU framework to bridge two supervisions, enabling partial setting to enjoy grounding bonuses of full supervision.

\begin{figure*}[t]
\begin{center}
\includegraphics[width=0.93\textwidth] {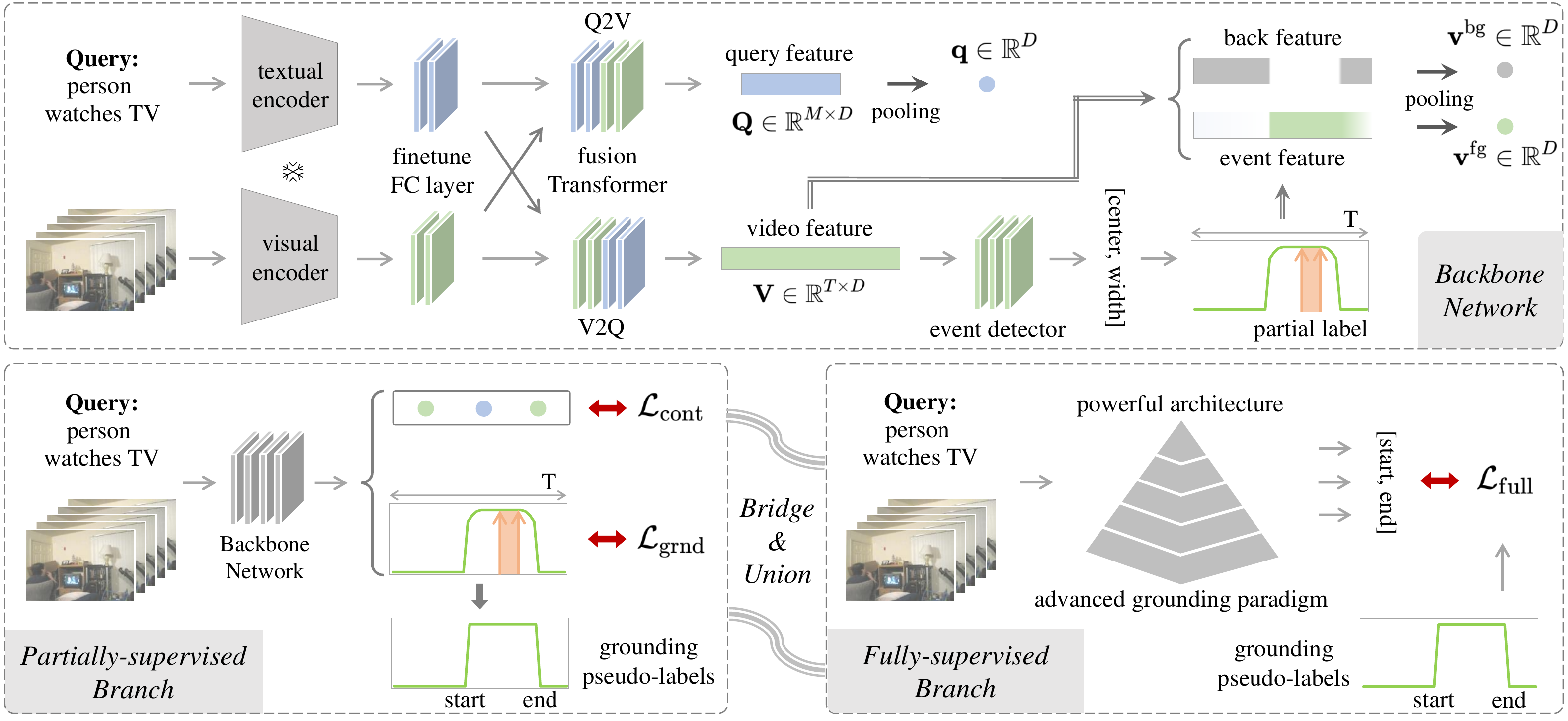}
\end{center}
\vspace{-0.4cm}
\caption{\textbf{Partial-Full Union Framework.} Given partial labels, we use one partially-supervised branch to facilitate high-quality alignment of event-query representations, {\em i.e.}, interact bi-modal features, adopt an event detector to calculate event visual features and background features, finally optimize with the constraint loss and partial grounding loss. Bridging by grounding pseudo-labels, we further unite the partially-supervised branch with a powerful fully-supervised branch, to enjoy the grounding bonus from fully-supervised research lines.}
\label{fig:framework}
\end{figure*}

\vspace{0.1cm}
{\noindent \bf Weakly-supervised Temporal Sentence Grounding} significantly alleviates annotation costs with only video-query pairs available. Its core is to learn high-quality multi-modal representations, including two branches. MIL-based methods~\cite{chen2020look,huang2021cross,lin2020weakly,mithun2019weakly,wang2021weakly,ju2021adaptive,ju2022distilling} follow multi-instance learning to generate video embeddings from redundant proposals, and used paired text embeddings for video-query contrastive learning. Reconstruction-based methods~\cite{lin2020weakly,song2020weakly,EC-SLweakly,zhang2020learning,TBVLweakly,zheng20221weakly,zheng20222weakly} reconstruct text queries using grounded information to optimize visual-textual embeddings and refine results. Despite the low annotation cost, weakly-supervised methods rest the poor performance, leaving a large gap from applications in reality. With partial labels, this paper greatly improves their performance at a slightly increased cost.

\vspace{0.1cm}
{\noindent \bf Partial Supervision} is designed to balance annotation costs and performance, including supervisions of click-level~\cite{papadopoulos2017training,lin2020interactive}, scribble-level~\cite{lin2016scribblesup,bearman2016s}, single-frame~\cite{ma2020sf,yang2021background}, or mixed-level~\cite{cheron2018flexible,biffi2020many}. Typically, in the image domain, several papers~\cite{bearman2016s,qian2019weakly,laradji2018blobs} proposed to click one point for each instance, and adopted metric learning between clicks for better visual representations.~\cite{lin2016scribblesup,bearman2016s} further improved click to more general scribble annotations. In the video domain,~\cite{mettes2016spot,moltisanti2019action} proposed single-point labels for spatial-temporal localization. Some papers explored single-frame labels~\cite{ju2021divide,ma2020sf,lee2021learning,ju2020point} or instance number labels~\cite{narayan20193c,xu2019segregated} for temporal detection. For temporal sentence grounding, ViGA~\cite{cui2022video} initialized single-frame settings. While in this paper, we introduce the more general partial supervision, by annotating short-clip (cover singe-frame) timestamps for each text query, to facilitate more cost-effective application in reality.

\section{Method}
\subsection{Formulation \& Preliminaries}
\noindent \textbf{Problem Formulation.}
Given an untrimmed video, Temporal Sentence Grounding (TSG) aims to design a model that detects the event boundary $(s, e)$ corresponding to the given text query, where $s \in \mathbb{R}$ and $e \in \mathbb{R}$ refer to the start timestamp and the end timestamp, respectively.

In this work, we consider the partially-supervised TSG setting (PTSG). For the $i$-th text query, only one short video clip $(t_i^s, t_i^e) \subseteq (s_i, e_i)$ is labeled by human annotators. We could also write it as $(t_i^c, r)$, where $t_i^c$ is the clip center and $r$ is the clip range. Note that, in the special case ($r=0$), this partial supervision degenerates to one single-frame setting~\cite{cui2022video}, that is, only $t_i \in [s_i, e_i]$ is annotated.

As a comparison, the fully-supervised setting annotates precise boundary $(s_i, e_i)$ for each query, while the weakly-supervised setting only has the prior of event existence.

\begin{figure}[t]
\begin{center}
\vspace{0.2cm}
\includegraphics[width=0.474\textwidth] {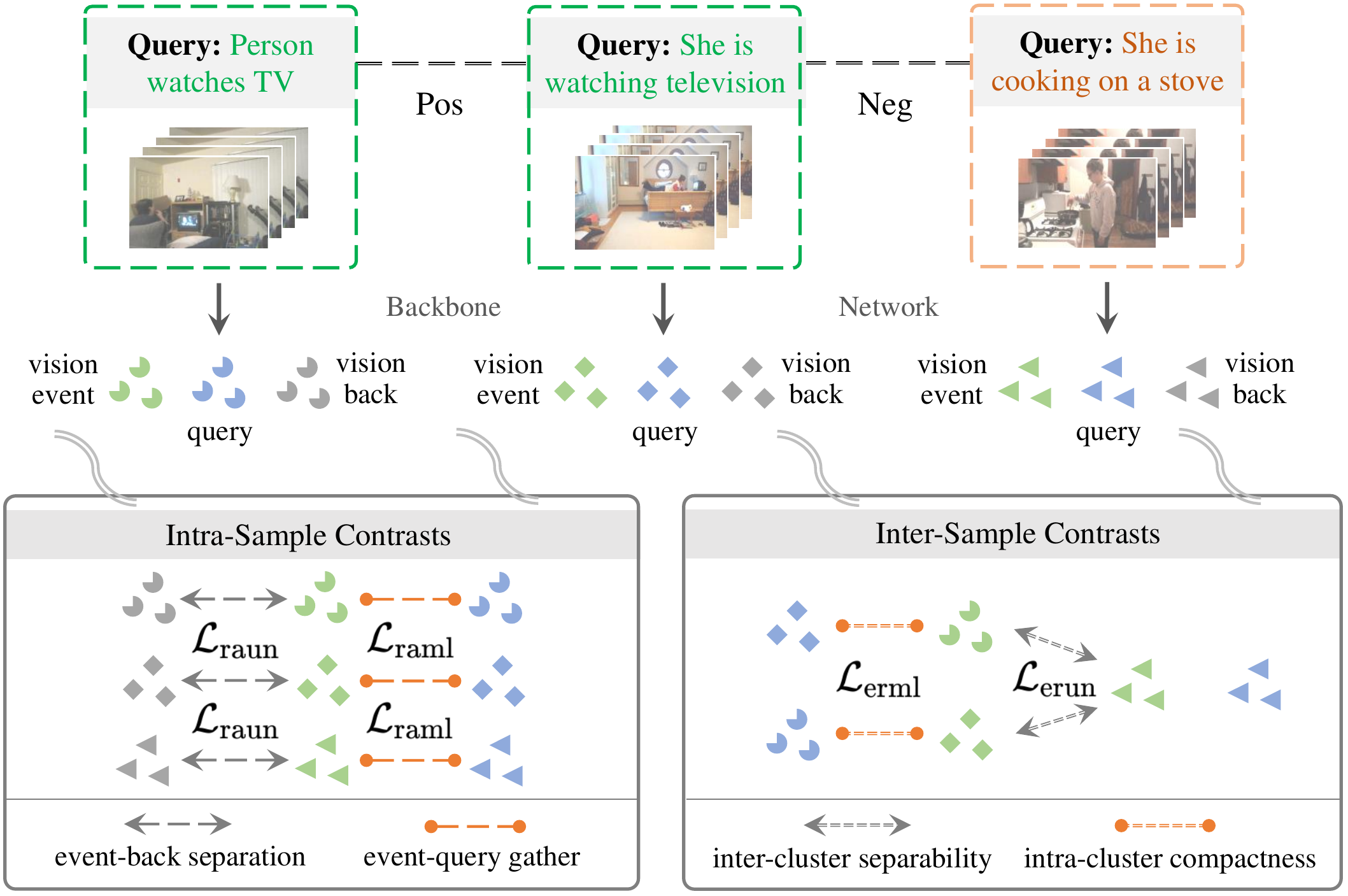}
\end{center}
\vspace{-0.4cm}
\caption{\textbf{Quadruple Constraint Pipeline.} We make full use of partial labels to align event-query representations with high quality. For intra-sample, $\mathcal{L}_{\mathrm{raml}}$ raises event-query gather, $\mathcal{L}_{\mathrm{raun}}$ enables event-background separation. For inter-sample, $\mathcal{L}_{\mathrm{erml}}$ and $\mathcal{L}_{\mathrm{erun}}$ urge intra-cluster compactness and inter-cluster separation.}
\label{fig:quadruple}
\end{figure}

\vspace{0.1cm}
\noindent \textbf{Motivation \& Overview.} To make full use of partial labels, we here design from: multi-modal representations and grounding framework. For high-quality alignment of event-query representations, we propose a novel quadruple constraint pipeline (Figure~\ref{fig:quadruple}), covering intra- and inter-sample, uni- and multi-modality constraints. For powerful performance with explicit grounding optimization, we introduce one partial-full union framework (Figure~\ref{fig:framework}), {\em i.e.}, generate pseudo-labels for an additional fully-supervised branch, to enjoy strong grounding bonuses from fully-supervised research lines, and be robust to partial annotations.

\vspace{0.1cm}
\noindent \textbf{Feature Extraction \& Fusion.}
Following existing methods~\cite{cui2022video,zheng20222weakly}, we pre-extract features for videos and queries, to save computing costs. For the video stream, we adopt the pre-trained 3D convolutional networks~\cite{tran2015learning,carreira2017quo}, and obtain $\mathbf{V'} \in \mathbb{R}^{T \times D_v}$; for the query stream, we adopt GloVe~\cite{pennington2014glove} to obtain $\mathbf{Q'} \in \mathbb{R}^{M \times D_q}$, where $T$, $M$, $D_v$, $D_q$ refer to the number of video frames, the number of query words, the video and query feature dimension respectively.

And then, we utilize one full-connection layer, to individually fine-tune these uni-modal features $\mathbf{V'}$ and $\mathbf{Q'}$, respectively. To interact bi-modal features, two cross-modal Transformers are further introduced, each by regarding one modality as the query, the other modality as key and value. Fused visual features are denoted as $\mathbf{V} \in \mathbb{R}^{T \times D}$, linguistic features as $\mathbf{Q} \in \mathbb{R}^{M \times D}$ ($D$ is the feature dimension).

\subsection{From Partial Label to Event Instance}   \label{sec:frame to intance}
With partial labels $(t^c, r)=(t^e,t^s)$, this section aims to detect the event interval of query $\mathbf{Q}$ from video $\mathbf{V}$.

\vspace{0.1cm}
\noindent \textbf{Event Detector.}
To represent the time interval for one event instance, there are two common solutions: frame-wise~\cite{cui2022video} and proposal-wise~\cite{zheng20222weakly}. The former continually determines whether each frame belongs to the text query, while the latter only regresses the event center and event width.

Compared with frame-wise, the proposal-wise solution has significant advantages under partial supervision. (1) It is parameter-efficient (two degrees of freedom for one event), greatly reducing the solution space; (2) It promotes temporal smoothness and rules out trivial predictions. Therefore, treating $t^c$ as the seed anchor, we here map the fused video features $\mathbf{V}$ to the center offset $\delta$ and event width $\ell$, through an event detector $\Phi(\cdot)$. And the corresponding start and end timestamps $[\widehat{s}, \, \widehat{e}]$ can be formulated as:
\begin{equation} \label{eq:instance}
\setlength{\abovedisplayskip}{7pt}
    {[\delta, \; \ell] = \Phi(\mathbf{V}),  \quad  \widehat{s} = p-\frac{\ell}{2},  \quad  \widehat{e} = p+\frac{\ell}{2},}
\setlength{\belowdisplayskip}{7pt}
\end{equation}
where $p=t^c+\delta$ means the center of the grounded event.

\vspace{0.1cm}
\noindent \textbf{Event Mask.}
Given the predicted start and end timestamps, we generate the temporal mask to facilitate the calculation of event-query representations. Here, we provide two solutions to convert $[\widehat{s}, \, \widehat{e}]$ into the differentiable mask $\mathbf{m} \in \mathbb{R}^{T}$. One is to approximate using the Gaussian-shape, and the other is to leverage the plateau-shape~\cite{moltisanti2019action}. Due to complete discrimination between event and background, the plateau-shaped mask outperforms the Gaussian-shaped mask (see the empirical comparisons in Table~\ref{tab:shape}). 

Using the resultant mask $\mathbf{m}$, we can filter out visual features for event $\mathbf{v}^{\mathrm{fg}} \in \mathbb{R}^{D}$, and background $\mathbf{v}^{\mathrm{bg}} \in \mathbb{R}^{D}$.
\begin{equation}
    {\mathbf{v}^{\mathrm{fg}} = \frac{1}{T}\sum_{t=1}^{T}{m_t}\mathbf{v}_t,
    \ \ \
    \mathbf{v}^{\mathrm{bg}} = \frac{1}{T}\sum_{t=1}^{T}{(1-m_t)}\mathbf{v}_t.
    }
\end{equation}

\noindent \textbf{Supervision.} Partial labels could provide limited yet precise supervision for the event mask $\mathbf{m}$, that is, the annotated short-clip is required to be included in the event interval:
\begin{equation}   \label{eq:partial}
{\mathcal{L}_{\mathrm{grnd}}=\mathrm{max}(t^e-\widehat{e}, \ \ \widehat{s}-t^s, \ 0),
}
\end{equation}
where $\mathrm{max}$ selects the maximum among the three.

\vspace{0.1cm}
\noindent \underline{\textit{Remark.}}
\hspace{2pt} Hereafter, we build constraints between (event, query) pairs, for the superior semantics and alignment than (video, query) or (short-clip, query) pairs (see Table~\ref{tab:representation}). We employ partial labels as the grounding anchors, which are essential to avoid false positives (see Table~\ref{tab:singleframe}).

\subsection{Quadruple Constraint Pipeline}
One key for TSG is to shape the high-quality embedding space, where visual-linguistic modalities are aligned. Here we perform a quadruple constraint pipeline to pursue comprehensive alignment, covering intra- and inter-sample, uni- and multi-modality, as shown in Figure~\ref{fig:quadruple}.

The quadruple constraint loss is calculated by:
\begin{align}
\mathcal{L}_{\mathrm{cont}} = (\mathcal{L}_{\mathrm{raml}} + \mathcal{L}_{\mathrm{raun}})  + \lambda(\mathcal{L}_{\mathrm{erml}} + \mathcal{L}_{\mathrm{erun}}),
\end{align}
where $\lambda$ refers to the balancing parameter.

\subsubsection{Intra-Sample Constraints}
We first consider the common solution, {\em i.e.}, constraint from the single sample. The available data is the visual features for event $\mathbf{v}^{\mathrm{fg}}$, and background $\mathbf{v}^{\mathrm{bg}}$. The query features $\mathbf{q} \in \mathbb{R}^{D}$ are obtained by mean-pooling word-wise features $\mathbf{Q}$.

\vspace{0.1cm}
\noindent \textbf{Event-Query Multi-Modal Constraint.} 
To enable paired event-query gather in the embedding space, we introduce the entire video $\mathbf{v}^{\mathrm{vd}}$ as a reference, by pooling the frame-wise video features $\mathbf{V}$. Since $\mathbf{v}^{\mathrm{vd}}$ contains both event and background, we promote the semantic similarity of ($\mathbf{v}^{\mathrm{fg}}$, $\mathbf{q}$) to be greater than that of ($\mathbf{v}^{\mathrm{vd}}$, $\mathbf{q}$): 
\begin{align}
\mathcal{L}_{\mathrm{raml}} = \mathrm{max}(\mathcal{S}(\mathbf{v}^{\mathrm{vd}}, \mathbf{q})-\mathcal{S}(\mathbf{v}^{\mathrm{fg}}, \mathbf{q})+\alpha, 0),
\end{align}
where $\mathcal{S}$ and $\alpha$ refer to the cosine similarity and the margin parameter. Such one constraint roughly obtains visual-linguistic correspondence, which is widely-adopted by existing methods from various levels of supervisions.

\vspace{0.1cm}
\noindent \textbf{Vision Uni-Modal Constraint.}
The videos contain richer fine-grained contexts than texts. Such contexts strengthen the continuity of videos, but result in similar features across events and backgrounds, making temporal grounding quite tricky. We here apply an additional visual-modal constraint, to raise better event-background separation.

Concretely, we also employ the entire video $\mathbf{v}^{\mathrm{vd}}$ for reference, {\em i.e.}, promote the semantic similarity of ($\mathbf{v}^{\mathrm{fg}}$, $\mathbf{v}^{\mathrm{vd}}$) to have more than $\beta$ margin than that of ($\mathbf{v}^{\mathrm{fg}}$, $\mathbf{v}^{\mathrm{bg}}$), in the embedding space, which can be formalized as:
\begin{align}
\mathcal{L}_{\mathrm{raun}} = \mathrm{max}(\mathcal{S}(\mathbf{v}^{\mathrm{fg}}, \mathbf{v}^{\mathrm{bg}})-\mathcal{S}(\mathbf{v}^{\mathrm{fg}}, \mathbf{v}^{\mathrm{vd}})+\beta, 0),
\end{align}

\subsubsection{Inter-Sample Constraints}
For representation learning, hard sample mining~\cite{dong2017class,han2020self} is essential to better shape multi-modal embedding space. Instead of naive data augmentation, we here mine the correlations between a large number of training samples, for superior inter-sample constraints.

For the vanilla classification tasks, inter-sample is readily available, since categories are clearly defined. However, for TSG, category clusters are not intuitive. To encourage data clusters, we consider using text queries as the bridge between samples for correlation establishing, given that language essentially refers to high-level semantics. To be specific, we calculate cosine similarity for all query features in the training set, and treat those with similarities exceeding the threshold $\theta$ as the same cluster. As a result, all samples can be clustered into $K$ clusters $\Lambda_1, \Lambda_2, \dots, \Lambda_K$. Another two contrast constraints are detailed to further enable intra-cluster compactness and inter-cluster separability.

\vspace{0.1cm}
\noindent \textbf{Event-Query Multi-Modal Constraint.}
In terms of multi-modal alignment, we regard the events and queries from the same cluster as the positive pairs, while those from different clusters as negative pairs. With this idea, denoting the positive set of ${i}$-th sample as $\Psi^{\mathrm{+}}_{i}$, and the negative set as $\Psi^{\mathrm{-}}_{i}$, inter-sample multi-modal contrast can be written as:
\begin{align}
\mathcal{L}_{\mathrm{erml}} = & \sum_i - \log \frac{\sum_{m \in {\Psi^{\mathrm{+}}_{i}}}  \exp(\mathbf{v}_i^{\mathrm{fg}} \cdot \mathbf{q}_{m}/ \tau)}{\sum_{j \in \{ \Psi^{\mathrm{+}}_{i} \cup \, \Psi^{\mathrm{-}}_{i} \}
} \exp(\mathbf{v}_i^{\mathrm{fg}} \cdot \mathbf{q}_{j}/ \tau)}.
\end{align}

\vspace{0.1cm}
\noindent \textbf{Vision Uni-Modal Constraint.}
In terms of modeling cluster semantics for the visual modality, we construct the positive set $\Omega^{\mathrm{+}}_{\mathrm{i}}$ by joining events from the same cluster of $i$-th sample, while build the negative set $\Omega^{\mathrm{-}}_{\mathrm{i}}$ by leveraging event features from the other clusters. That is,
\begin{align}
\mathcal{L}_{\mathrm{erun}} = & \sum_i - \log \frac{\sum_{m \in {\Omega^{\mathrm{+}}_{\mathrm{i}}}}  \exp(\mathbf{v}_i^{\mathrm{fg}} \cdot \mathbf{v}_{m}^{\mathrm{fg}} / \tau)}{\sum_{j \in \{ \Omega^{\mathrm{+}}_{\mathrm{i}} \cup \, \Omega^{\mathrm{-}}_{\mathrm{i}} \}
} \exp(\mathbf{v}_i^{\mathrm{fg}} \cdot \mathbf{v}_{j}^{\mathrm{fg}}  / \tau)}.
\end{align}

\noindent \underline{\textit{Remark.}}
\hspace{1pt} For intra-sample constraints, we learn at event-query level with suitable semantics, than short-clip-query or video-query levels. For inter-sample constraints, we adopt text queries to bridge data of similar semantics. As videos contain abundant redundant contexts, these clustered samples could act as hard positive or negative samples for each other, which enables better representations than trivial data augmentation (see comparisons in Table~\ref{tab:representation}).

\subsection{Partial-Full Union Framework}
In this section, we leverage the above powerful representations to generate temporal grounding results. To achieve the goal, one trivial solution is to output predictions from the event detector (in Eq.~\ref{eq:instance}). However, it suffers from several issues. (1) Partial labels, which provide grounding anchors to avoid false positives, are unavailable for inference; (2) Lacking explicit optimization for grounding objectives, leading to excessive noise in results or pseudo-labels.

To tackle these issues, we introduce a partial-full union (PFU) framework, by uniting an additional fully-supervised branch to the partial-supervised branch. To be specific, we provide the learned representations for both branches, while utilize the pseudo-labels generated from Eq.~\ref{eq:instance} of the partial branch, as the grounding objectives for the full branch.

Although this pipeline is simple, its role is to do more at one stroke, and the insight contained is non-trivial.

\textit{Structurally}, this framework bridges full and partial supervisions, which enables the partial setting to enjoy the superior grounding bonus from the existing fully-supervised methods. For example, explicit grounding optimization objectives, better pyramid architecture~\cite{li2021proposal}, and advanced detection paradigm~\cite{li2022compositional}. \textit{Functionally}, this framework plays a similar role as self-training or knowledge distillation. The fully-supervised branch could further denoise and refine the pseudo-labels from partially-supervised branch to achieve better grounding results. From \textit{an annotation perspective}, this framework avoids the labeling gap between training and testing for partial supervision, by only using the fully-supervised branch for efficient inference. Moreover, partial labels have high degree of freedom in event intervals, which poses a great challenge to the robustness of the framework. With the help of this full-supervised branch, our PFU framework could achieve consistent effectiveness in various label distributions. Furthermore, our framework is flexible to handle two levels of supervisions, enabling to jointly learn from wider data, may bring stronger performance.

\vspace{0.1cm}
\noindent \underline{\textit{Discussion.}}
Our PFU framework bridges the gap of PTSG and FTSG, meaning that all existing fully-supervised methods can be leveraged as our full branch. Table~\ref{tab:fullsupervision} demonstrates that the stronger method brings better results for our framework. As more advanced methods become available, our performance is promising to improve continuously.

\input{tab/SOTA.tex}

\input{tab/Single-Frame.tex}

\subsection{Training and Inference}
To train the partially-supervised branch, we balance the constraint loss $\mathcal{L}_{\mathrm{cont}}$ and grounding loss $\mathcal{L}_{\mathrm{grnd}}$ by $\gamma$. 
\begin{align}
\mathcal{L}_{\mathrm{part}} = \mathcal{L}_{\mathrm{cont}} + \gamma\mathcal{L}_{\mathrm{grnd}}.
\end{align}

To train the fully-supervised branch, we follow~\cite{liu2021progressively,rodriguez2020proposal,jiang2022sdn} and denote the optimization loss as $\mathcal{L}_{\mathrm{full}}$. 

At testing time, we utilize the grounding results from the fully-supervised branch directly, for efficient inference.

\section{Experiments}
\subsection{Implementation}
\noindent \textbf{Datasets.} \textbf{Charades-STA} has 9,848 videos recording daily indoors activities, which can be divided into 5338 training videos of 12,408 moment-query annotations, and 1334 testing videos of 3720 moment-query pairs. \textbf{ActivityNet Captions} contains 19,994 videos of diverse domains. Following the convention, 37,421 and 17,505 moment-query annotations are provided for training and validating, respectively. 17,031 moment-query pairs are used for evaluation.

\vspace{0.1cm}
\noindent \textbf{Metrics.}
Following existing works~\cite{cui2022video,zheng20222weakly}, we here evaluate the results through `R@K, IoU=M', {\em i.e.}, the percentage of predicted moments with Intersection over Union (IoU) greater than M in the top-K recall. To evaluate the quality of grounding pseudo-labels, we also report mean Intersection over Union (mIoU) over ground-truth labels.

\vspace{0.1cm}
\noindent \textbf{Details.}
Our framework is implemented with Pytorch, using Adam with a learning rate of $0.0004$, and a batch size of $32$ samples. For network architectures, two cross-model Transformers are identical, both with $3$ layers of $4$ heads for encoder and decoder. For inter-sample clustering, we first obtain query features using Bert~\cite{reimers2019sentence}, then cluster samples with similarities over threshold $\theta=0.8$. All hyperparameters are set by the grid search: $\lambda=0.5$, $\tau=0.13$, $\gamma=3$. On Charades-STA, $\alpha=0.25$ and $\beta=0.32$; while on ActivityNet Captions, $\alpha=0.18$ and $\beta=0.18$.

\subsection{Comparison with State-of-the-art}
In this section, we perform comprehensive comparisons with current state-of-the-art methods from various levels of supervision, to prove the effectiveness of our method.

\vspace{0.1cm}
\noindent \textbf{Single-frame supervision.}
Table~\ref{tab:SOTA} demonstrates the comparison across multiple IoU thresholds on both datasets. For the sake of fairness, we utilize the identical single-frame annotations following~\cite{cui2022video}. Generally speaking, our framework achieves new state-of-the-art under most IoU regimes. For example, 4.08\% mIoU gains over the previous SOTA on Charades-STA, narrowing the performance gap between full and partial supervisions by a large margin. Moreover, our method gains more on the rigorous evaluation than loose regimes, {\em e.g.}, 0.36\% gains for R@0.3 \textit{vs.} 8.07\% gains for R@0.7, comparing to ViGA~\cite{cui2022video}. As rigorous regimes are more practical in reality, our grounding results are proved to be more complete and precise. Despite being partial supervision, at the low IoU regimes, our method is even comparable with some earlier fully-supervised methods~\cite{zhang2020span,mun2020local}, again showing the gratifying performance.

Note that, ActivityNet Captions is challenging even for fully-supervised methods, resulting in a small performance gap between various supervisions on this dataset. Overall, our method is comparable to the SOTA~\cite{cui2022video}.

\vspace{0.1cm}
\noindent \textbf{Short-clip supervision.}
For partial supervision, short-clip annotations are also economical. We also evaluate their effectiveness in Table~\ref{tab:SOTA}, and here each short-clip is annotated for 2 seconds. On both datasets, adding this small range of timestamps bring around 1\% mIoU gains over single-frame labels, at an acceptable cost of annotation time, further narrowing the performance gap with full supervision.

\input{tab/Ablation.tex}

\input{tab/Sentence.tex}

\subsection{Ablation Study \& Discussion}
In this section, we conduct thorough ablation studies to dissect all key components, both quantitatively and qualitatively. Unless otherwise stated, we all perform experiments using single-frame annotations on Charades-STA.

\vspace{0.1cm}
\noindent \textbf{Framework Robustness.} 
Partial labels have a high degree of freedom in the event intervals, which poses great challenges to framework robustness. Table~\ref{tab:singleframe} simulates single-frame labels with multiple samplings of two distributions, {\em i.e.}, Uniform and Gaussian; annotates short-clip labels for various durations, to evaluate the pseudo-label quality from partially-supervised branch. Our framework shows consistent effectiveness to various annotations and distributions, proving strong robustness. Comparing to random sampling, Gaussian can cause labels closer to the event center, which eases the grounding difficulty and brings better results, but is impractical. For short-clip labels, longer durations lead to better results yet greater annotation burdens.

In Eq.~\ref{eq:partial}, we exploit partial annotations by treating them as precise grounding seed anchors, and provide loss $\mathcal{L}_{\mathrm{grnd}}$. To demonstrate the effectiveness, we remove this constraint for single-frame labels, and report the results in `Weak' of Table~\ref{tab:singleframe}. Such the partial grounding loss limits the grounding interval of the framework, thus largely avoiding false-positive predictions, and improving pseudo-labels.

\input{tab/Fullsupervision.tex}

\input{tab/Shape.tex}

\vspace{0.1cm}
\noindent \textbf{Contribution of quadruple constraints.}
To achieve great representations, we design quadruple constraints: for intra-sample, uni-modal loss $\mathcal{L}_{\mathrm{raun}}$ and multi-modal loss $\mathcal{L}_{\mathrm{raml}}$; for inter-sample, uni-modal loss $\mathcal{L}_{\mathrm{erun}}$ and multi-modal loss $\mathcal{L}_{\mathrm{erml}}$. As shown in Table~\ref{tab:ablation}, the single $\mathcal{L}_{\mathrm{raml}}$ (A1) causes poor pseudo-labels for the partial branch. A2 adds $\mathcal{L}_{\mathrm{raun}}$ to encourage event-back separation, thus obtaining clear improvements. Happily, inter-sample modeling brings immediate gains. By introducing $\mathcal{L}_{\mathrm{erml}}$ to A1, A3 gets more than 13\% mIoU gains; by adding $\mathcal{L}_{\mathrm{erun}}$ to A3, A5 further gets 1.1\% mIoU gains, showing the advantages of sample relationship modeling. Note that, comparing to the uni-modal losses, multi-modal losses usually bring more gains, because multi-modal contrasts also implicitly promote event-background separation. In conclusion, all losses are essential and jointly contribute to the best results.

\vspace{0.1cm}
\noindent \textbf{Effectiveness of representation learning.}
Comparing to the constraints between (video, query) pairs or (short-clip, query) pairs, we propose (event, query) aligned pairs. Table~\ref{tab:representation} evaluates the effect of these representations, under the identical settings. As is evident, the event-query representation outperforms the other two by a large margin, obtaining 4.4\% mIoU gains over the video-query representation. Since it only considers partial video-clips as positives for queries, `short-clip-query' wrongly inhibits the completeness of grounding, resulting in bad performance.

\begin{figure}[t]
\begin{center}
\includegraphics[width=0.46\textwidth] {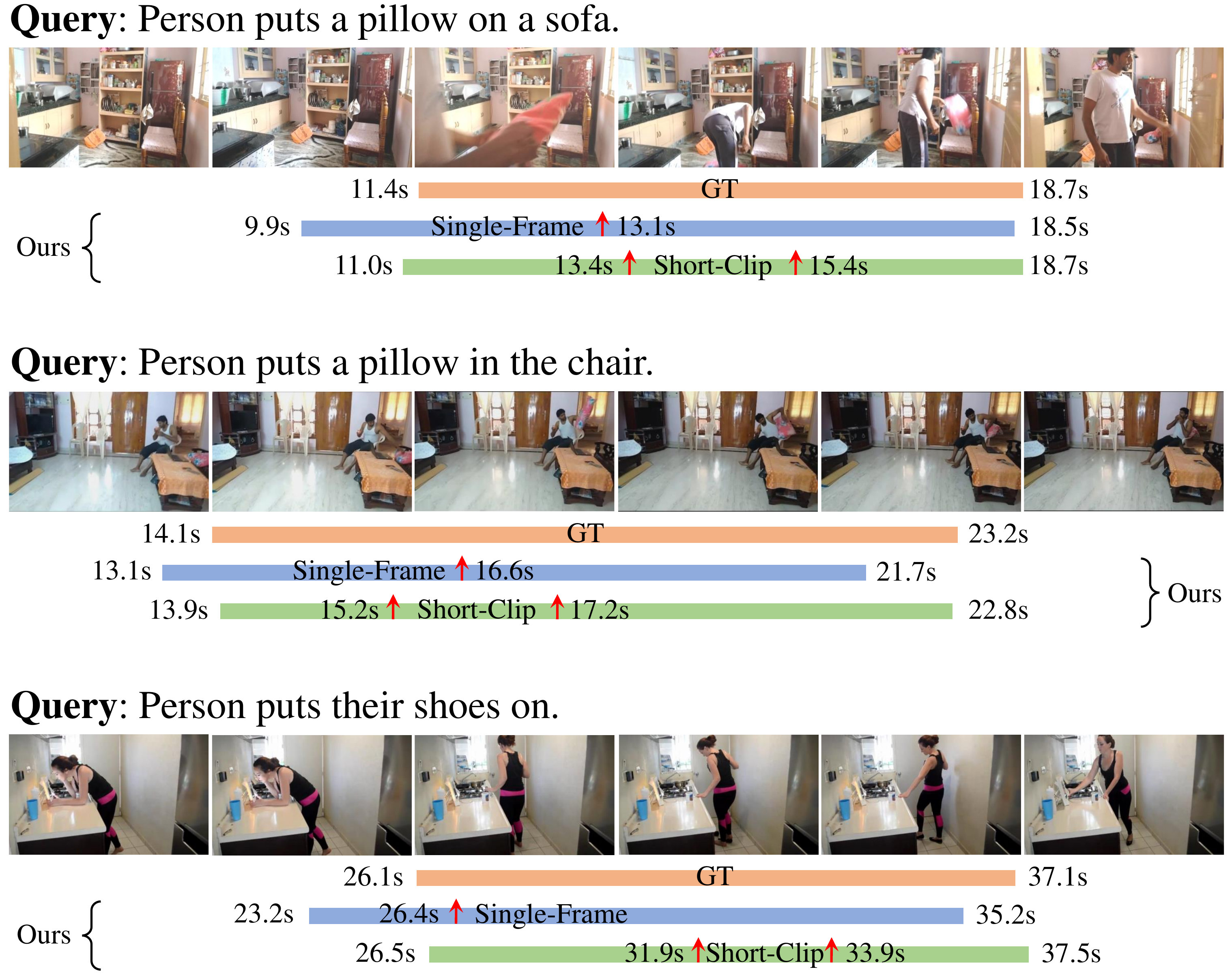}
\end{center}
\vspace{-0.5cm}
\caption{\textbf{Visualization.} Using partial labels as seed anchors, our method obtains great pseudo-labels from the partially-supervised branch. Short-clip labels further improve single-frame results.}
\label{fig:results}
\end{figure}

\vspace{0.1cm}
\noindent \textbf{Effectiveness of inter-sample constraints.}
Table~\ref{tab:representation} reveals the efficacy of inter-sample constraints. In terms of clustering for cross-sample relationships, we construct a baseline: randomly augment the video as its own positive. Comparing B3 to B1, we conclude that it is effective to identify semantic consistency of videos using the query similarity. And it is efficient to mine hard samples than simple data augmentation. Besides that, we also reveal the effectiveness of inter-video negative samples by comparing B3 to B2. Consistent with self-supervised learning~\cite{he2020momentum,caron2020unsupervised,chen2020big}, without negative samples, the framework falls into the trivial solution, {\em i.e.}, collapsing all representations together.

\vspace{0.1cm}
\noindent \textbf{Generalization of the PFU framework.}
Our framework bridges the gap of PTSG and FTSG, thus can jointly process data from two supervisions, and employ existing fully-supervised methods for help. Table~\ref{tab:fullsupervision} evaluates its generalization. Three typical fully-supervised methods, namely, IA-Net~\cite{liu2021progressively}, TMLGA~\cite{rodriguez2020proposal}, and SDN~\cite{jiang2022sdn} are reproduced (through official source codes) for comparison. Although annotations vary greatly, our partially-supervised method just performs slightly worse than fully-supervised counterparts. Such small performance gaps reveal the good generalization. In addition, the stronger fully-supervised method leads to better results for our method. With more advanced methods become available (open source), our performance is promising to be continuously improved in the future.

\vspace{0.1cm}
\noindent \textbf{Comparison of mask shapes.}
Table~\ref{tab:shape} makes comparisons for two solutions in the event mask (section~\ref{sec:frame to intance}), {\em i.e.}, utilizing the Gaussian function or plateau function. For a clearer analysis, we also report three results for the plateau function under different steepness of side slopes. The plateau-shape mask is superior to the Gaussian-shaped mask, and the performance becomes better as plateau-shape closer to the gate-shape, demonstrating the importance of completely distinguishing event features and background features.

\begin{figure}[t]
\begin{center}
\includegraphics[width=0.46\textwidth] {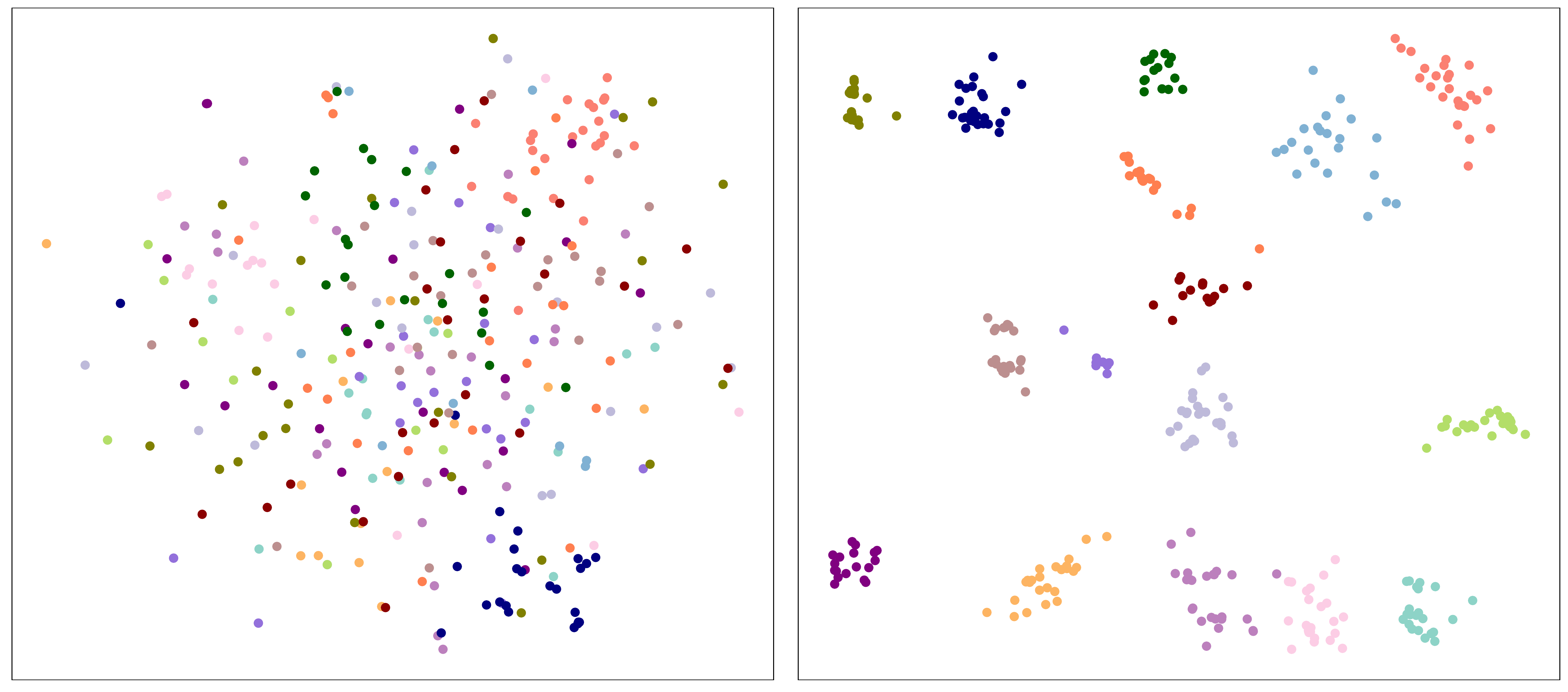}
\end{center}
\vspace{-0.4cm}
\caption{\small \textbf{Representation T-SNE.} Left are original video features from pre-trained extractors. Right are video features with quadruple constraint pipeline. Colors indicate different clusters.}
\label{fig:tsne}
\end{figure}

\subsection{Qualitative Analysis}
\noindent \textbf{Grounding Results.}
To intuitively demonstrate our superiority, Figure~\ref{fig:results} visualizes several grounding pseudo-labels from the partially-supervised branch, giving both single-frame labels and short-clip (2 seconds) labels. As is evident, in all three cases of different scenarios, our method outputs high-quality pseudo-labels, using partial labels as anchors. Comparing to single-frame labels, short-clip labels provide more guidance, thus further correcting results to be better.

\vspace{0.1cm}
\noindent \textbf{Representation Results.}
To attain further insights into the learned representations, we collect some videos, and visualize their features using T-SNE in Figure~\ref{fig:tsne}. Different semantic clusters are distinguished by colors. In general, original video features (left) from pre-trained extractors completely confuse different clusters. Such chaotic semantic relationships usually lead to low-quality grounding results. While after adopting the quadruple constraint pipeline, video representations show good inter-cluster separability and intra-cluster compactness, facilitating better grounding.

\section{Conclusion}
This paper proposes partial supervision for TSG to balance performance and annotation burdens. To make full use of such limited yet precise grounding information, we design separately for multi-modal representations and grounding framework. For the former, we propose the quadruple constraint pipeline, consisting of comprehensive constraints from intra-sample, inter-sample, uni-modality, and multi-modality, to align the event-query features with high quality. For the latter, we bridge the partially-supervised branch with a fully-supervised branch by grounding pseudo-labels, thus turning PTSG into FTSG to share the powerful grounding bonus from fully-supervised methods. Extensive experiments and thorough ablations demonstrate the significance of partial supervision, and our superior performance over existing methods on two public datasets.

\vspace{0.2cm}
{\small
\bibliographystyle{ieee_fullname}
\bibliography{egbib}
}

\clearpage
\twocolumn

\begin{appendices}
{
  \hypersetup{linkcolor=black}
}

\section{Implementation Details}
\subsection{Training of the Partially-supervised Branch}
Our partial branch is implemented with PyTorch, and all experiments are done on one 24G GeForce RTX 3090 GPU. On both datasets, the model is optimized with Adam~\cite{kingma2014adam} optimizer using a learning rate of $0.0004$, and a batch size of $32$ samples. We train this branch for $30$ epochs to ensure better convergence, under all partial annotations.

Given one sample, we pre-extract features for videos and queries to save computing costs, following existing methods~\cite{cui2022video,zheng20222weakly}. For the visual stream, we first downsample the video to a fixed FPS of $8$, then perform feature extraction using the I3D network~\cite{carreira2017quo} ($D_v=1024$) for Charades-STA, the C3D network~\cite{tran2015learning} ($D_v=500$) for ActivityNet Captions. To deal with the large variety in video durations, we also pad each video with all zeros to $T=200$ frames on both datasets. For the textual stream, we adopt GloVe~\cite{pennington2014glove} to extract word-wise embeddings ($D_q=300$). The maximum word number $M=20$, the vocabulary size is $8000$.

For inter-sample relationship modeling across the entire dataset, we leverage the pre-trained Transformer Bert~\cite{reimers2019sentence}, a powerful sentence encoder, to extract the query features. We calculate the cosine similarity between any two query features, and then regard those whose similarity exceeds the threshold $\theta$ as the same semantic cluster. Within one batch, we randomly sample eight semantic clusters, each of which contains four (video, query) samples for training.

\subsection{Training of the Fully-supervised Branch}
To enjoy the grounding bonus from the fully-supervised research line, we here train one additional fully-supervised branch, using grounding pseudo-labels from the partially-supervised branch. Concretely, we adopt three typical fully-supervised methods~\cite{liu2021progressively,rodriguez2020proposal,jiang2022sdn} for experiment. We follow their own hyperparameters and designs, just replacing ground-truth with pseudo-labels for training. The grounding results are reported in Table~\ref{tab:fullsupervision} of the main paper.

\subsection{Inference Details}
During testing, we directly utilize the grounding results from the fully-supervised branch for the efficient inference. Taking SDN~\cite{jiang2022sdn} for an example, given one video, it predicts the frame-level boundary (start \& end) probability and frame-level binary (event or background) probability for each query. Then along the temporal dimension, it selects the boundary timestamps with the max start or end probability as the event proposal, and scores each proposal by the average binary probability. Please refer to~\cite{liu2021progressively,rodriguez2020proposal,jiang2022sdn} for more detailed architectures, designs, and optimizations.

To evaluate pseudo-labels and grounding results, we employ `R@1, IoU=M' and mIoU, following the literature. That is, the percentage of predicted moments with Intersection over Union (IoU) greater than the threshold M in the Top-1 recall. For simplicity, we abbreviate `R@1, IoU=M' to `R@M' across all Tables of the main paper.

\begin{figure*}[t]
\begin{center}
\includegraphics[width=0.985\textwidth] {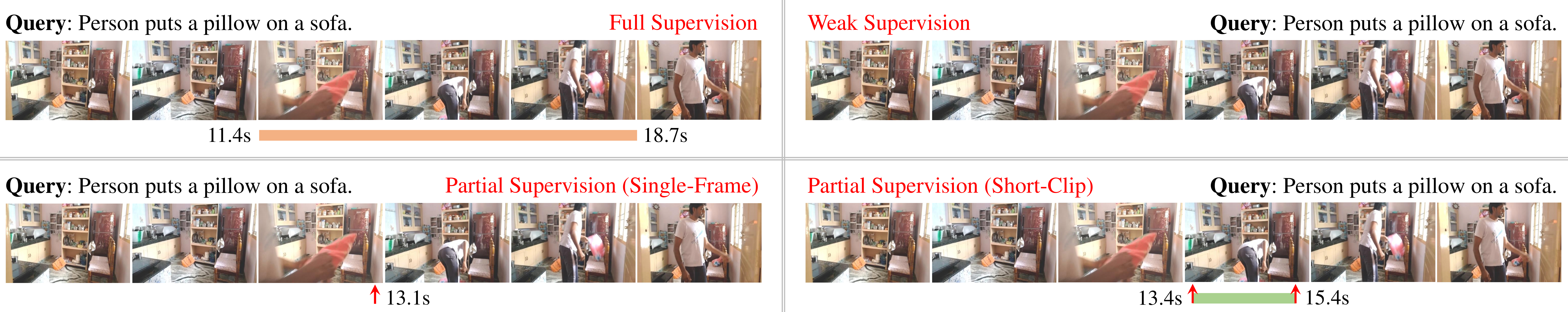}
\end{center}
\vspace{-0.5cm}
\caption{\textbf{Supervision Settings.} During training, full supervision provides precise boundary labels for each query; weak supervision only annotates the video-query pair; our partial supervision annotates one short video clip (or even single frame in the strictest case) for each query. During testing, only video-query pairs are available for all supervisions, without any grounding labels.}
\vspace{0.2cm}
\label{fig:compare}
\end{figure*}

\section{Partial Supervision}
\subsection{Datasets}
\textbf{Charades-STA} consists of 9,848 videos recording daily indoors activities, which can be divided into 5338 training videos of 12,408 event-query annotations, and 1334 testing videos of 3720 event-query pairs. 

\textbf{ActivityNet Captions} overs 19,994 videos from diverse domains. As conventions, 37,421 and 17,505 event-query annotations are provided for training and validating. While 17,031 event-query pairs are used for evaluation.

\subsection{Partial Annotations}
In this paper, we consider the partially-supervised setting for temporal sentence grounding (TSG). Figure~\ref{fig:compare} illustrates the differences between various supervision settings.

During training, full supervision provides precise boundaries (start \& end timestamps) $(s_i, e_i)$ for each query. Weak supervision only provides the information on video-query pairs. For our proposed partial supervision, one short video clip $(t_i^s, t_i^e) \subseteq (s_i, e_i)$ is randomly labeled for each query by human annotators. In the special case ($t_i^s=t_i^e$), this partial supervision degenerates to one single-frame setting~\cite{cui2022video}, that is, only $t_i \in [s_i, e_i]$ is randomly annotated.

During testing, only video-query pairs are available for all supervisions, without any grounding labels. Note that, in our partially-supervised branch, we regard partial labels as grounding anchors, and extend them for event intervals. The labeling gap between training and testing makes evaluation via only partially-supervised branch infeasible. We bridge a fully-supervised branch (PFU) to solve this issue.

\subsection{Labeling Procedure}
Before annotating, we provide several (video, query) examples for the annotator, to familiarize him with the dataset. During annotating, given one (video, query) pair, the annotator is asked to pause the video when he identifies the event corresponding to the query. And this pause timestamp is regarded as one single-frame annotation. Then the annotator needs to continue watching the video for a few seconds from the pause timestamp, forming one short-clip annotation.

While watching, the annotator is able to quickly browse, pause and go to any timestamp. Comparing to single-frame labels, the short-clip labels only add a few seconds of time costs per video, so they are also very efficient.

\subsection{Performance \,vs. Annotation Costs}
In Figure~\ref{fig:intro}, we evaluate the performance in terms of annotation costs between various supervisions, on Charades-STA. Due to the large variety in video durations, we adopt annotation time/video duration to measure annotation costs. The average results across all videos are reported.

Full supervision has good performance, but requires expensive annotation costs, as the annotator usually watches the video many times to define the boundary timestamps. Weak supervision significantly alleviates annotation costs, but achieves poor performance. The annotation cost for partial labels is close to weak supervision, but much fewer than full supervision. The performance of partial labels is close to full supervision, far exceeding weak supervision.

\section{Limitations and Future Work}
To save computing resources, we freeze the pre-trained feature extractors during training, which could potentially bias to the raw pre-training data. Moreover, some state-of-the-art methods in full supervision are not open source up to now, which also somewhat limits our performance.

As the future work, we expect more computing resources available, making it possible to end-to-end optimize the proposed framework using all training data. In addition, with more advanced fully-supervised methods becoming available (or open source), our partially-supervised performance is promising to be further improved.
\end{appendices}

\end{document}

%% file: tab/SOTA.tex
\begin{table*}[t]
\footnotesize
\setlength\tabcolsep{6.5pt}
\centering
\begin{tabular}{cc|c|ccccc|ccccc}
\toprule 
\multicolumn{2}{c|}{\multirow{2}{*}{Supervision}} & \multirow{2}{*}{Method} & \multicolumn{5}{c|}{Charades-STA} & \multicolumn{5}{c}{ActivityNet Captions} \\ \cline{4-13} 
\multicolumn{2}{c|}{} &  & R@0.1 & R@0.3 & R@0.5 & R@0.7 & mIoU & R@0.1 & R@0.3 & R@0.5 & R@0.7 & mIoU \\ \hline
\multicolumn{2}{c|}{\multirow{7}{*}{Full}} & D-TSG~\cite{liu2022reducing} & -- & -- & 65.05 & 42.77 & \multicolumn{1}{c|}{--} & -- & -- & 54.29 & 33.64 & -- \\
\multicolumn{2}{c|}{} & MGSL~\cite{liu2022memory} & -- & -- & 63.98 & 41.03 & \multicolumn{1}{c|}{--} & -- & -- & 51.87 & 31.42 & -- \\
\multicolumn{2}{c|}{} & CBLN~\cite{liu2021context} & -- & -- & 61.13 & 38.22 & \multicolumn{1}{c|}{--} & -- & 66.34 & 48.12 & 27.60 & -- \\
\multicolumn{2}{c|}{} & SDN~\cite{jiang2022sdn} & 82.72 & 73.71 & 59.89 & 41.80 & \multicolumn{1}{c|}{54.13} & 84.75 & 60.88 & 42.03 & 26.36 & 43.38 \\
\multicolumn{2}{c|}{} & LGI~\cite{mun2020local} & -- & 72.96 & 59.46 & 35.48 & \multicolumn{1}{c|}{51.38} & -- & 58.52 & 41.51 & 23.07 & 41.13 \\
\multicolumn{2}{c|}{} & VSLNet~\cite{zhang2020span} & -- & 70.46 & 54.19 & 35.22 & \multicolumn{1}{c|}{50.02} & -- & 63.16 & 43.22 & 26.16 & 43.19 \\
\multicolumn{2}{c|}{} & TMLGA~\cite{rodriguez2020proposal} & 81.59 & 69.62 & 50.11 & 32.50 & \multicolumn{1}{c|}{48.28} & 75.25 & 51.28 & 33.04 & 19.26 & 37.78 \\
\hline \hline
\multicolumn{2}{c|}{\multirow{5}{*}{Weak}} & LCNet~\cite{yang2021local} & -- & 59.60 & 39.19 & 18.87 & \multicolumn{1}{c|}{--} & 78.58 & 48.49 & 26.33 & -- & -- \\
\multicolumn{2}{c|}{} & VCA~\cite{wang2021visual} & -- & 58.58 & 38.13 & 19.57 & \multicolumn{1}{c|}{--} & 67.96 & 50.45 & 31.00 & -- & -- \\
\multicolumn{2}{c|}{} & CRM~\cite{huang2021cross} & - & 53.66 & 34.76 & 16.37 & \multicolumn{1}{c|}{--} & 81.61 & 55.26 & 32.19 & -- & -- \\
\multicolumn{2}{c|}{} & CNM~\cite{zheng20221weakly} & -- & 60.39 & 35.43 & 15.45 & \multicolumn{1}{c|}{--} & 78.13 & 55.68 & 33.33 & -- & -- \\
\multicolumn{2}{c|}{} & CPL~\cite{zheng20222weakly} & 75.93 & 67.07 & 48.83 & 22.61 & \multicolumn{1}{c|}{43.71} & 82.18 & 55.28 & 30.61 & 12.32 & 36.82 \\
\hline \hline
\multicolumn{1}{c|}{\multirow{5}{*}{Partial}} & \multirow{4}{*}{\begin{tabular}[c]{@{}c@{}}Single-\\ Frame\end{tabular}} &  2D-TAN~\cite{zhang2020learning} & -- & -- & -- & -- & \multicolumn{1}{c|}{--} & -- & 11.26 & 5.28 & 2.34 & -- \\
\multicolumn{1}{c|}{} &  & LGI~\cite{mun2020local} & -- & 51.94 & 25.67 & 7.98 & \multicolumn{1}{c|}{30.83} & -- & 9.34 & 4.11 & 1.31 & 7.82 \\
\multicolumn{1}{c|}{} &  & ViGA~\cite{cui2022video} & -- & 71.21 & 45.05 & 20.27 & \multicolumn{1}{c|}{44.57} & -- & {59.61} & {35.79} & {\bf 16.96} & {40.12} \\
\multicolumn{1}{c|}{} &  & \textbf{Ours} & {\bf 82.10} & {\bf 71.57} & {\bf 54.66} & {\bf 28.34} & \multicolumn{1}{c|}{{\bf 48.65}} & {\bf 83.45} & {\bf 59.63} & {\bf 36.35} & {16.61} & {\bf 40.15} \\  \cline{2-13} 
\multicolumn{1}{c|}{} & Short-Clip & \textbf{Ours} & {\bf 82.42} & {\bf 72.09} & {\bf 56.43} & {\bf 32.08} & \multicolumn{1}{c|}{{\bf 49.91}} & {\bf 84.70} & {\bf 60.55} & {\bf 36.84} & {\bf 16.98} & {\bf 40.73} \\ 
\bottomrule
\end{tabular}
\vspace{-0.1cm}
\caption{\textbf{Comparison with State-of-the-art.} Using single-frame labels, our method outperforms most competitors by a large margin. The performance can be further improved with short-clip labels (2 seconds), and even comparable to several earlier fully-supervised methods.}
\label{tab:SOTA}
\end{table*}

%% file: tab/Single-Frame.tex
\begin{table}[t]
\footnotesize
\setlength\tabcolsep{7pt}
\centering
\vspace{0.2cm}
\begin{tabular}{c|c|cccc}
\toprule
Setting & Distribution & R@0.3 & R@0.5 & R@0.7 & mIoU \\ \hline  \hline
Weak & -- & 46.47 & 22.33 & 9.16 & 28.24 \\ \hline \hline
\multirow{4}{*}{\begin{tabular}[c]{@{}c@{}}Single\\ Frame\end{tabular}}
 & Uniform-1 & 97.77 & 71.32 & 28.59 & 60.50 \\
 & Uniform-2 & 97.25 & 71.50 & 28.94 & 60.35 \\
 & Uniform-3 & 97.71 & 72.04 & 29.48 & 60.69 \\ \cline{2-6} 
 & Gaussian & 98.23 & 78.08 & 34.70 & 63.17 \\  \hline \hline
\multirow{3}{*}{\begin{tabular}[c]{@{}c@{}}Short\\ Clip\end{tabular}} 
 & 2-seconds & 99.69 & 89.05 & 40.46 & 67.01 \\
 & 3-seconds & 99.84 & 89.94 & 44.52 & 68.07 \\
 & 4-seconds & 99.89 & 92.79 & 52.76 & 70.56 \\
\bottomrule
\end{tabular}
\vspace{-0.1cm}
\caption{\textbf{Effectiveness of partial annotations.} Weak refers to not using the partial grounding loss $\mathcal{L}_{\mathrm{grnd}}$ (Eq.~\ref{eq:partial}). For single-frame or short-clip labels from various types of labeling distributions, our framework demonstrates strong robustness and superiority.}
\label{tab:singleframe}
\end{table}

%% file: tab/Ablation.tex
\begin{table}[t]
\footnotesize
\setlength\tabcolsep{4.2pt}
\centering
\begin{tabular}{c|cccc|ccc}
\toprule 
& $\mathcal{L}_{\mathrm{raml}}$ & $\mathcal{L}_{\mathrm{raun}}$ & $\mathcal{L}_{\mathrm{erml}}$ & $\mathcal{L}_{\mathrm{erun}}$ & R@0.5 & R@0.7 & mIoU \\ 
\hline \hline
A1 & \checkmark & &  &  & 31.78 & 8.91 & 44.30 \\
A2 & \checkmark & \checkmark &  &  & 36.41 & 10.63 & 46.41 \\
A3 & \checkmark &  & \checkmark &  & 62.99 & 25.90 & 57.67 \\
A4 & \checkmark & \checkmark & \checkmark &  & 67.64 & 27.52 & 58.90 \\
A5 & \checkmark & & \checkmark & \checkmark & 67.40 & 27.58 & 58.71 \\
A6 & \checkmark & \checkmark & \checkmark & \checkmark & 71.32 & 28.59 & 60.50 \\ 
\bottomrule
\end{tabular}
\vspace{-0.1cm}
\caption{\textbf{Ablation study of quadruple constraint pipeline.} We evaluate the quality of pseudo-labels under single-frame annotations.
All losses jointly contribute to the best performance.}
\vspace{0.3cm}
\label{tab:ablation}
\end{table}

%% file: tab/Sentence.tex
\begin{table}[t]
\footnotesize
\setlength\tabcolsep{5.2pt}
\centering
\begin{tabular}{c|c|cc|cc}
\toprule 
\multirow{2}{*}{} & \multirow{2}{*}{Representation} & \multicolumn{2}{c|}{Inter-Sample} & \multirow{2}{*}{R@0.7} & \multirow{2}{*}{mIoU} \\ \cline{3-4}
   &  & Relevance & Sample &  &    \\ \hline \hline
B1 & \multirow{3}{*}{(event, query)} & Augment & Pos+Neg & 20.81 & 54.02  \\
B2 &  & Similar & Pos & 3.06 &  35.32  \\
B3 &  & Similar & Pos+Neg & 28.59 & 60.50   \\   \hline
B4 & (video, query) & Similar & Pos+Neg & 21.01 &  56.11  \\
B5 & (short-clip, query) & Similar & Pos+Neg & 18.67 & 51.39  \\ 
\bottomrule
\end{tabular}
\vspace{-0.1cm}
\caption{\textbf{Comparison of constraint designs.} For representation learning, the event-query pairs are significantly superior to video-query or short-clip-query pairs. For the inter-sample modeling, our relevance mining using similar queries brings great gains, comparing to vanilla data augmentation. `Pos' and `Neg' refer to mining for positive samples and negative samples, respectively.}
\label{tab:representation}
\end{table}

%% file: tab/Fullsupervision.tex
\begin{table}[t]
\footnotesize
\setlength\tabcolsep{6.2pt}
\centering
\begin{tabular}{c|c|cccc}
\toprule 
Method & Label & R@0.3 & R@0.5 & R@0.7 & mIoU \\ \hline  \hline
\multirow{3}{*}{\begin{tabular}[c]{@{}c@{}}IA-Net\\ \cite{liu2021progressively}\end{tabular}} & Full & 68.87 & 57.00 & 28.27 & 46.63 \\
 & Single-Frame & 65.36 & 53.60 & 25.87 & 44.70 \\
 & Short-Clip & 67.22 & 55.52 & 27.41 & 45.74 \\\hline \hline
\multirow{3}{*}{\begin{tabular}[c]{@{}c@{}}TMLGA\\ \cite{rodriguez2020proposal}\end{tabular}} & Full & 69.62 & 50.11 & 32.50 & 48.28 \\
 & Single-Frame & 67.61 & 45.08 & 26.34 & 45.36 \\
 & Short-Clip & 67.26 & 50.97 & 28.55 & 46.10 \\\hline \hline
\multirow{3}{*}{\begin{tabular}[c]{@{}c@{}}SDN\\ \cite{jiang2022sdn}\end{tabular}}
 & Full & 73.71 & 59.89 & 41.80 & 54.13 \\
 & Single-Frame & 71.57 & 54.66 & 28.34 & 48.65 \\ 
 & Short-Clip & 72.09 & 56.43 & 32.08 & 49.91 \\ 
\bottomrule
\end{tabular}
\vspace{-0.1cm}
\caption{\textbf{Generalization of partial-full union framework.} Our method can deal with data from full or partial supervision. Bridging our partial branch to various fully-supervised methods brings promising results. Stronger methods lead to better performance.}
\vspace{0.3cm}
\label{tab:fullsupervision}
\end{table}

%% file: tab/Shape.tex
\begin{table}[t]
\footnotesize
\setlength\tabcolsep{8pt}
\centering
\begin{tabular}{c|c|cccc}
\toprule 
Shape & Steep & R@0.3 & R@0.5 & R@0.7 & mIoU \\ \hline  \hline
Gaussian & -- & 91.21 & 54.53 & 21.63 & 54.12 \\ \hline \hline
\multirow{3}{*}{Plateau} 
 & 20 & 95.58 & 56.87 & 20.28 & 55.42 \\
 & 40 & 96.22 & 69.54 & 28.57 & 59.68 \\
 & 60 & 97.77 & 71.32 & 28.59 & 60.50 \\ 
\bottomrule
\end{tabular}
\vspace{-0.1cm}
\caption{\textbf{Comparison of mask shapes.} Plateau-shape clearly distinguishes the event from the background, thus being better than Gaussian-shape. Steep refers to the steepness of side slopes.}
\label{tab:shape}
\end{table}